\documentclass[runningheads,a4paper]{llncs}

\usepackage{amssymb}
\usepackage{graphicx}
\usepackage{tabularx}
\usepackage{multirow}
\usepackage[hidelinks]{hyperref}
\usepackage{url}
\usepackage[dvipsnames]{xcolor}
\urldef{\mailsa}\path|{alfred.hofmann, ursula.barth, ingrid.haas, frank.holzwarth,|
\urldef{\mailsb}\path|anna.kramer, leonie.kunz, christine.reiss, nicole.sator,|
\urldef{\mailsc}\path|erika.siebert-cole, peter.strasser, lncs}@springer.com|    
\newcommand{\keywords}[1]{\par\addvspace\baselineskip
\noindent\keywordname\enspace\ignorespaces#1}
\begin{document}

\mainmatter  

\title{{\scriptsize Presented at \textit{NVIDIA GTC, The Conference for the Era of AI and the Metaverse}, \\March 23, 2023. [S51129]} \\ \textcolor{white}{.}
 \\
Evaluating XGBoost for Balanced and Imbalanced Data\\
Application to Fraud Detection}
\titlerunning{Evaluating XGBoost for Balanced and Imbalanced Data}

%
%
\author{Gissel Velarde$^*$, Anindya Sudhir, Sanjay Deshmane, Anuj Deshmunkh, Khushboo Sharma, Vaibhav Joshi}
\authorrunning{Evaluating XGBoost for Balanced and Imbalanced Data}

\institute{Vodafone,\\
Ferdinand Platz 1, Germany\\
$^*$gissel.velarde@vodafone.com\\
\url{http://www.vodafone.com}}

%
%

\toctitle{Evaluating XGBoost for Balanced and Imbalanced Data}
\tocauthor{Application to Fraud Detection}
\maketitle

\begin{abstract}
This paper evaluates XGboost's performance given different dataset sizes and class distributions, from perfectly balanced to highly imbalanced. XGBoost has been selected for evaluation, as it stands out in several benchmarks due to its detection performance and speed. After introducing the problem of fraud detection, the paper reviews evaluation metrics for detection systems or binary classifiers, and illustrates with examples how different metrics work for balanced and imbalanced datasets. Then, it examines the principles of XGBoost. It proposes a pipeline for data preparation and compares a Vanilla XGBoost against a random search-tuned XGBoost. Random search fine-tuning provides consistent improvement for large datasets of 100 thousand samples, not so for medium and small datasets of 10 and 1 thousand samples, respectively. Besides, as expected, XGBoost recognition performance improves as more data is available, and deteriorates detection performance as the datasets become more imbalanced. Tests on distributions with 50, 45, 25, and 5 percent positive samples show that the largest drop in detection performance occurs for the distribution with only 5 percent positive samples. Sampling to balance the training set does not provide consistent improvement. Therefore, future work will include a systematic study of different techniques to deal with data imbalance and evaluating other approaches, including graphs, autoencoders, and generative adversarial methods, to deal with the lack of labels.

\end{abstract}

\section{Introduction} \label{s:introduction}
Classification is a widely applied machine learning task in industrial setups. Outside laboratories, there might be few cases where class distribution is balanced, since most real-world problems deal with imbalanced datasets. Binary classification systems are evaluated on their ability to correctly identify negative and positive samples. Often, detecting positive samples is critical. 

In fraud detection, positive class samples may represent substantial business losses. At the same time, negative samples are essential, and therefore, flagging a negative sample as positive, is a lost business opportunity. Furthermore, the challenges are the following:

\begin{itemize}
\item fraudsters continuously change their behavior,
\item they may represent rare cases,
\item fraud patterns may even be unseen during training, and 
\item there might be a considerable delay until fraud is identified.
\end{itemize}

In 2021, estimations in the telecommunications sector report that loss due to fraud accounts for USD 39.89 Billion, representing over two percent of the global revenue of USD 1.8 Trillion \cite{Jacob2021}. From the several types of fraud, we are most concerned about equipment theft, commissions fraud, and device reselling, in which global losses were estimated USD 3.11 Billion, USD 2.15 Billion, and USD 1.67 Billion, respectively \cite{Jacob2021}.

In recent years, eXtreme Gradient Boosting (XGBoost) has gained attention since it proved highly competitive in machine learning contests for its recognition performance and speed. In this study, XGBoost is systematically evaluated on small, medium, and large datasets presenting different class distributions. In the first experiment, XGboost is evaluated given perfectly balanced data up to highly imbalanced data, where the positive cases represent only 5 percent of all samples. 

The contributions of this paper are the following:
\begin{itemize}
\item It provides examples to illustrate how different evaluation measures are to be interpreted for detection systems or binary classifiers.
\item It reviews the principles of Boosting Trees and the advantages of XGBoost as the selected boosting system.
\item It explains a pipeline for a Vanilla XGBoost and a random search-tuned XGBoost.  
\item It demonstrates empirically that XGboost performance increases its detection performance as the dataset size increases.
\item It shows that XGBoost's performance decreases as the data becomes more imbalanced.
\item It tests sampling to balance training set to deal with data imbalance.  
\end{itemize}

The following section reviews evaluation in binary detection systems or binary classifiers. Section \ref{s:xgboost}, reviews XGBoost. The experimental section can be found in section \ref{s:experiments}. Finally, conclusions are drawn in section \ref{s:conclusions}. 

\section{Evaluation in Detection Systems} \label{s:evaluation}
\begin{figure}
\centering
\includegraphics[height=8cm]{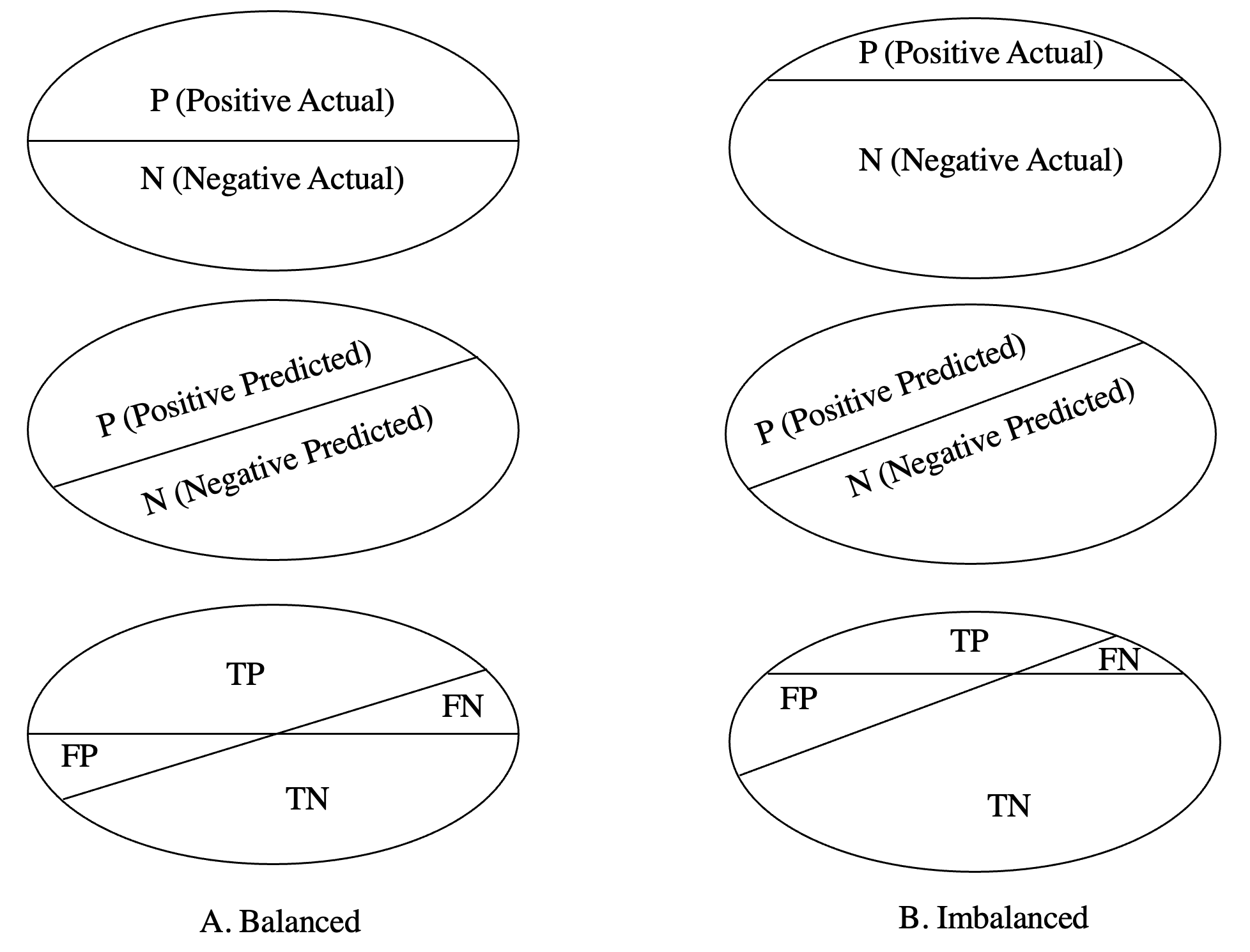}
\caption{Examples of possible distributions for balanced and imbalanced datasets.}
\label{fig:balanced_imbalanced}
\end{figure}

In detection systems or binary classifiers, we deal with Negative (N) and
Positive (P) samples, where the total number of samples is equal to $N + P$. Detection systems are evaluated considering detection performance, observing the number of True Positive (TP), True Negatives (TN), False Positives (FN), and False Negatives (FN) samples. See Fig. \ref{fig:balanced_imbalanced}. 

Detection systems or classifiers are evaluated considering the following \cite{saito2015precision},\cite{chicco2021matthews}: 
\begin{itemize}
\item Confusion Matrix,
\item Area Under Precision-Recall Curve (AUPRC) also known as Precision/Recall Curve (PRC).
\item Precision@n,
\item $F$ scores, depending on the case $F_1$, $F_{0.5}$ or $F_2$,
\item Matthews Correlation Coefficient (MCC),
\item False Positive Rate, False Negative Rate,
\item Revenue, or Costs, and
\item Execution time, among other measures.
\end{itemize}

\begin{table}[]
\begin{center}
\caption{Confusion Matrix} \label{t:cm}
\begin{tabular}{llll}
                                                               &                                 & \multicolumn{2}{c}{\textbf{Actual Class}}                                           \\ \cline{2-4} 
\multicolumn{1}{l|}{\textbf{}}                                 & \multicolumn{1}{l|}{\textbf{}}  & \multicolumn{1}{l|}{\textbf{P}}          & \multicolumn{1}{l|}{\textbf{N}}          \\ \cline{2-4} 
\multicolumn{1}{l|}{\multirow{2}{*}{\textbf{Predicted Class}}} & \multicolumn{1}{l|}{\textbf{P}} & \multicolumn{1}{l|}{True Positive (TP)}  & \multicolumn{1}{l|}{False Positive (FP)} \\ \cline{2-4} 
\multicolumn{1}{l|}{}                                          & \multicolumn{1}{l|}{\textbf{N}} & \multicolumn{1}{l|}{False Negative (FN)} & \multicolumn{1}{l|}{True Negative (TN)}  \\ \cline{2-4} 
\end{tabular}
\end{center}
\end{table}

In this paper, we will focus on the Confusion Matrix, Precision, Recall, and $F$ scores as relevant measures to evaluate detection systems. Although, Receiver Operating Characteristic (ROC) curve, is still widely used for evaluation, it is only a powerful tool for balanced datasets, and not recommended for imbalanced datasets \cite{saito2015precision}. Likewise, 
 Accuracy is deceiving when the dataset is imbalanced. Examples are provided later on.

The confusion matrix shown in Table \ref{t:cm} allows us to compute  \cite{saito2015precision}:
\begin{equation}
Baseline PRC = \frac{P}{P+N},
\end{equation}

\begin{equation}
Precision = \frac{TP}{TP+FP},
\end{equation}

\begin{equation}
Recall = \frac{TP}{TP+FN},
\end{equation}

\begin{equation}
F_\beta = (1+\beta^{2})\cdot \frac{Precision \cdot Recall}{\beta^{2}\cdot Precision+Recall},
\end{equation}
where $F_1$ gives same weight to Precision and Recall, $F_{0.5}$ gives more weight to Precision, and $F_{2}$ gives more weight to Recall.

In addition, we can compute:
\begin{equation}
Accuracy = \frac{TP+TN}{TP+FP+TN+FN},
\end{equation}

however, as mentioned before, Accuracy is not recommended when datasets are imbalanced. For instance, see Table \ref{t:1}, which presents five examples with 1000 samples each. Example 1 has an equal number of Positive and Negative samples, 500 each, respectively. Examples 2 to 5 have 900 Negative and 100 Positive samples. The Baseline PRC is 0.50 for Example 1 and 0.10 for Examples 2 to 5. 

\subsubsection{Examples 1 and 2.} The classifiers make an equal number of mistakes for FP and FN, such that Precision, Recall, $F_1$, $F_{0.5}$, $F_{2}$ are equal to $0.50$. Accuracy is $0.67$ for Example 1 and $0.90$ for Example 2, although the classifier in the second example still makes the same amount of mistakes, that is, the same number of FP and FN. 

\subsubsection{Examples 3.} It showcases a classifier that flags everything as Negative. In this case, only Recall and Accuracy can be computed, and again Accuracy gives a misleading score of 0.90. 

\subsubsection{Examples 4.} The classifier in this example flags everything as Positive. In this case, Recall is 1. The rest of the measures reflect better the detection ability of such a classifier. As expected, $F_{0.5}$ is worse than $F_1$ and $F_{2}$, because $F_{0.5}$ gives more weight to Recall. 

\subsubsection{Examples 5.} It showcases a classifier with high Precision but low Recall. Because this classifier makes no mistakes, it is highly precise, but it identifies only five Positive samples out of 100, and therefore its Recall is low. $F$ scores behave as expected. %

\begin{table}[]
\caption{Performance for five detection systems (classifiers) when a dataset is balanced (Example 1) and imbalanced (Examples 2 to 5).} \label{t:1}
\includegraphics[width=1\textwidth]{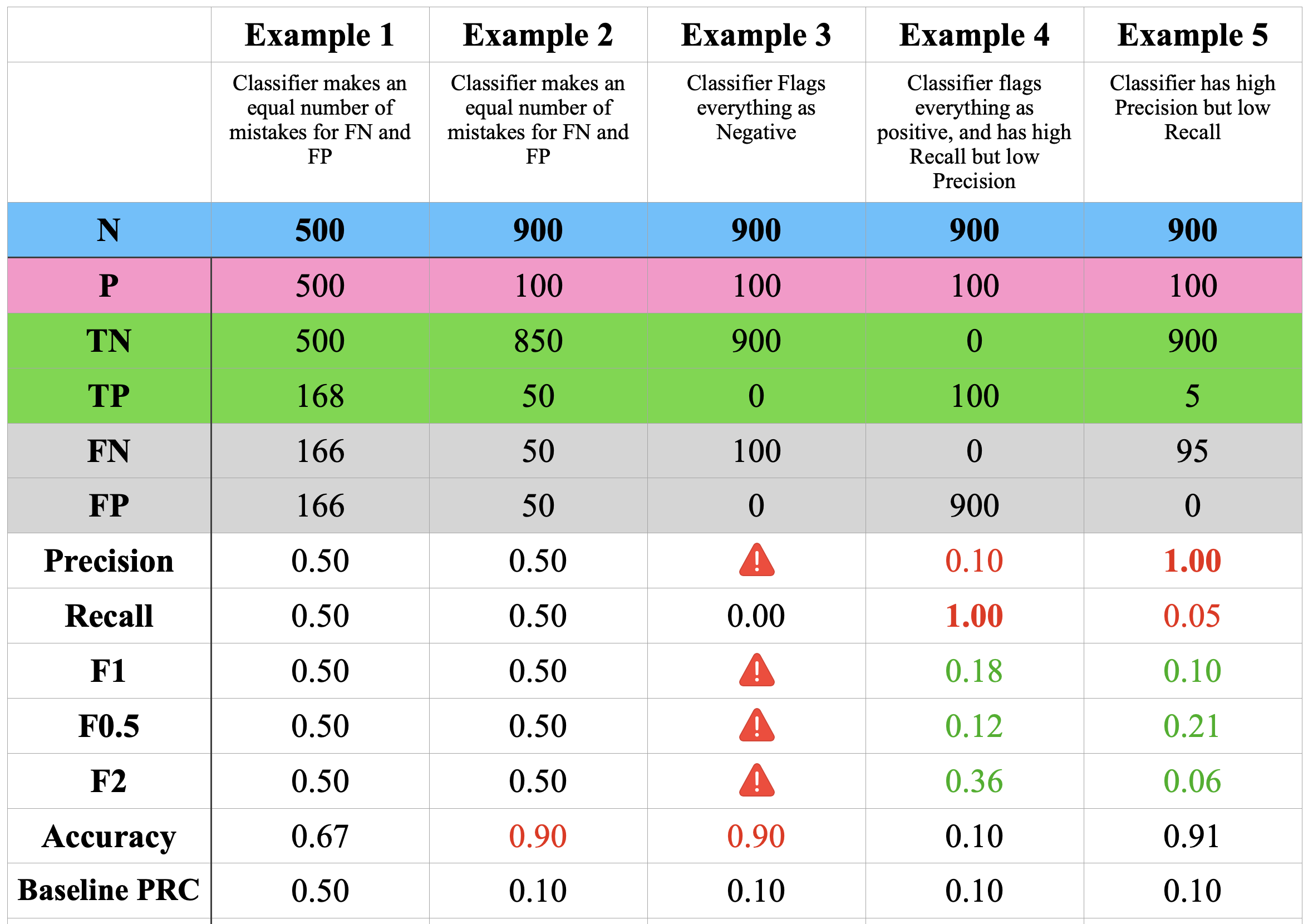}
\end{table}

The previous examples teach us that while it is important to look at Precision and Recall separately, $F$ scores summarise the performance of classifiers. In addition, Accuracy is deceiving when datasets are imbalanced, and therefore, is not recommended for evaluation.

Likewise, the ROC curve, although being a widely used tool to evaluate detection systems or classifiers, is a strong tool only when datasets are balanced and misleading when they are imbalanced \cite{saito2015precision}.

\section{XGBoost Review} \label{s:xgboost}
\begin{figure}[]
\caption{Example of a tree ensemble model with two trees. Decision nodes are oval and leaf nodes are rectangular. Each tree contributes to the final prediction. } \label{f:exampleTree}
\includegraphics[width=1\textwidth]{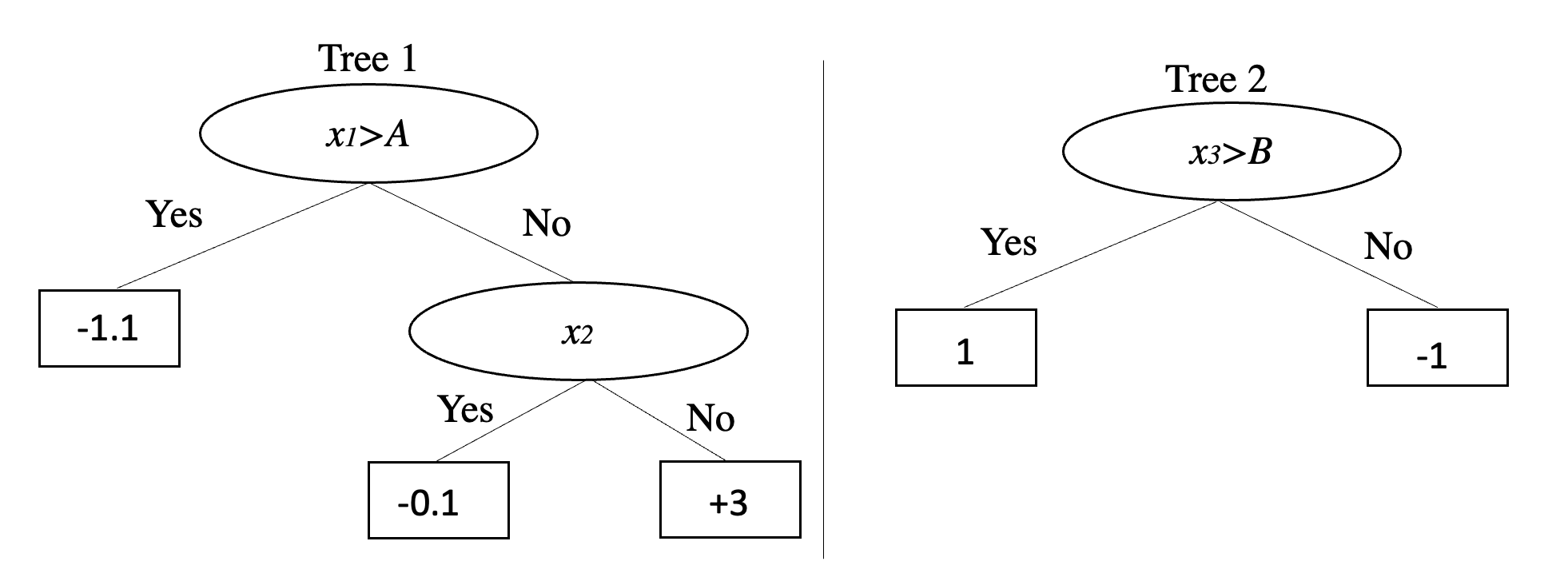}
\end{figure}

Extreme Gradient Boosting (XGBoost) \cite{chen2016xgboost} is a powerful tree boosting method \cite{friedman2001greedy} that: 
\begin{itemize}
\item First, creates a decision tree,
\item Then, iterates over $M$ number of trees, such that
\begin{itemize}
	\item	It builds a tree likely selecting samples that were misclassified by the previous tree.
\end{itemize}
\end{itemize}

See Figure \ref{f:exampleTree}. On tabular data, XGBoost has been reported to win several machine learning competitions \cite{chen2016xgboost}. Related work shows that XGBoost outperforms several machine learning algorithms for fraud detection in mobile payment \cite{hajek2022fraud}. The authors used a synthetic dataset called \textit{Paysim} containing more than 6 million samples with nine attributes or features, where only 0.13 percent of samples are positive. The study evaluated several supervised and unsupervised learning algorithms, reporting that Random Forest (RF) obtains the second best $F_1$ for supervised learning algorithms, being XGBoost much faster than RF. Besides, XGBOD, which is presented as a semi-supervised learning method performs best, but to train XGBOD, labels are required just like in a supervised setting, and XGBOD is very slow compared to XGBoost. Multi-Objective Generative Adversarial Active Learning (MO-GAAL) returns the fourth best score and it is the slowest of all algorithms. See Table \ref{t:2}. 

\begin{figure}[]
\caption{Comparison of tree boosting systems considering time. Taken from \protect\cite{chen2016xgboost}.} \label{f:s2}
\includegraphics[width=1\textwidth]{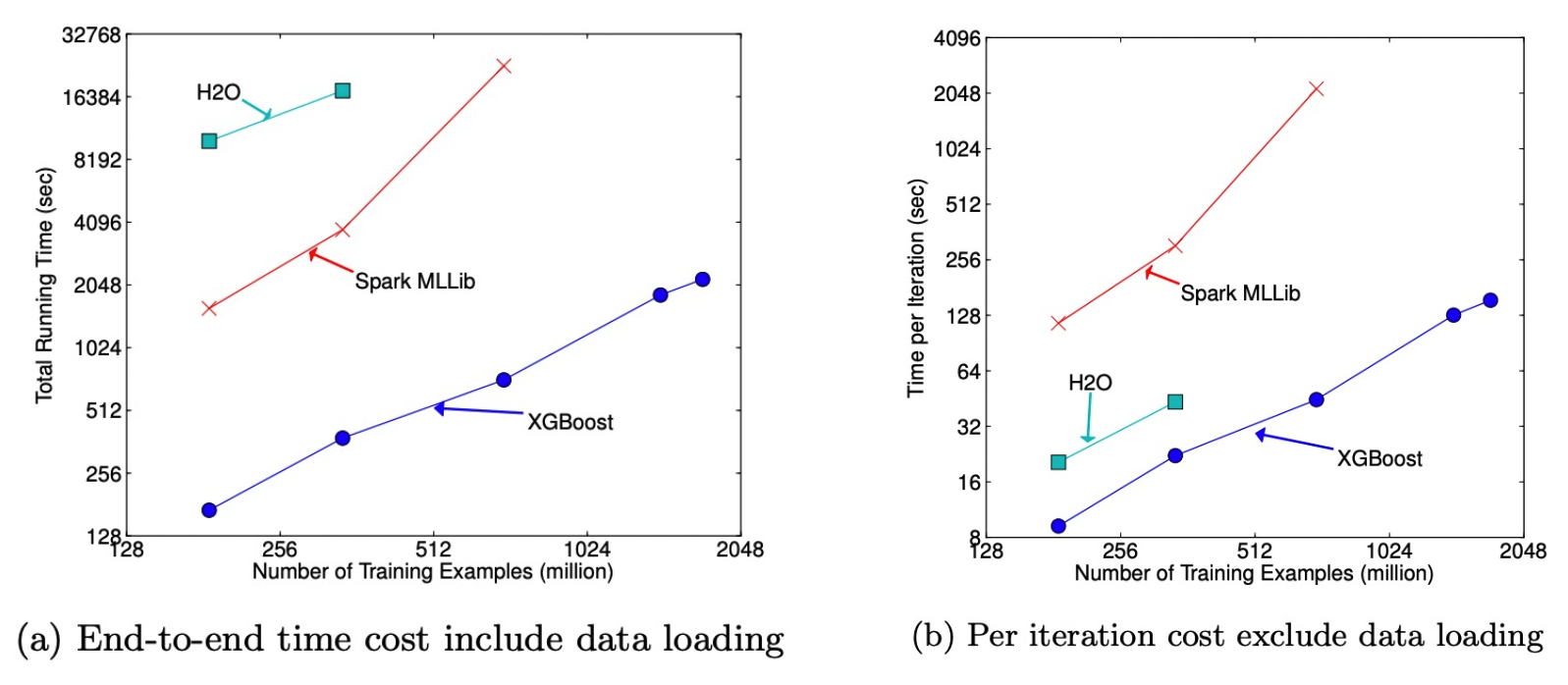}
\end{figure}

Table \ref{f:s1} and Figure \ref{f:s2} present comparisons between XGBoost and other boosting systems \cite{chen2016xgboost}. XGBoost is the fastest and possesses several characteristics, like sparsity awareness and parallel computing, that made it our choice for experimentation. The Exact Greedy Algorithm deals with finding the bests splits and enumerates all of them over all features, being computationally expensive. Therefore, XGBoost implements approximate local (per split) and global (per tree) solutions to the problem of finding the best splits \cite{chen2016xgboost}.

\begin{table}[]
\caption{Detection performance for various supervised and unsupervised methods on \textit{Paysim} dataset. Taken from \protect\cite{hajek2022fraud}.} \label{t:2}
\includegraphics[width=1\textwidth]{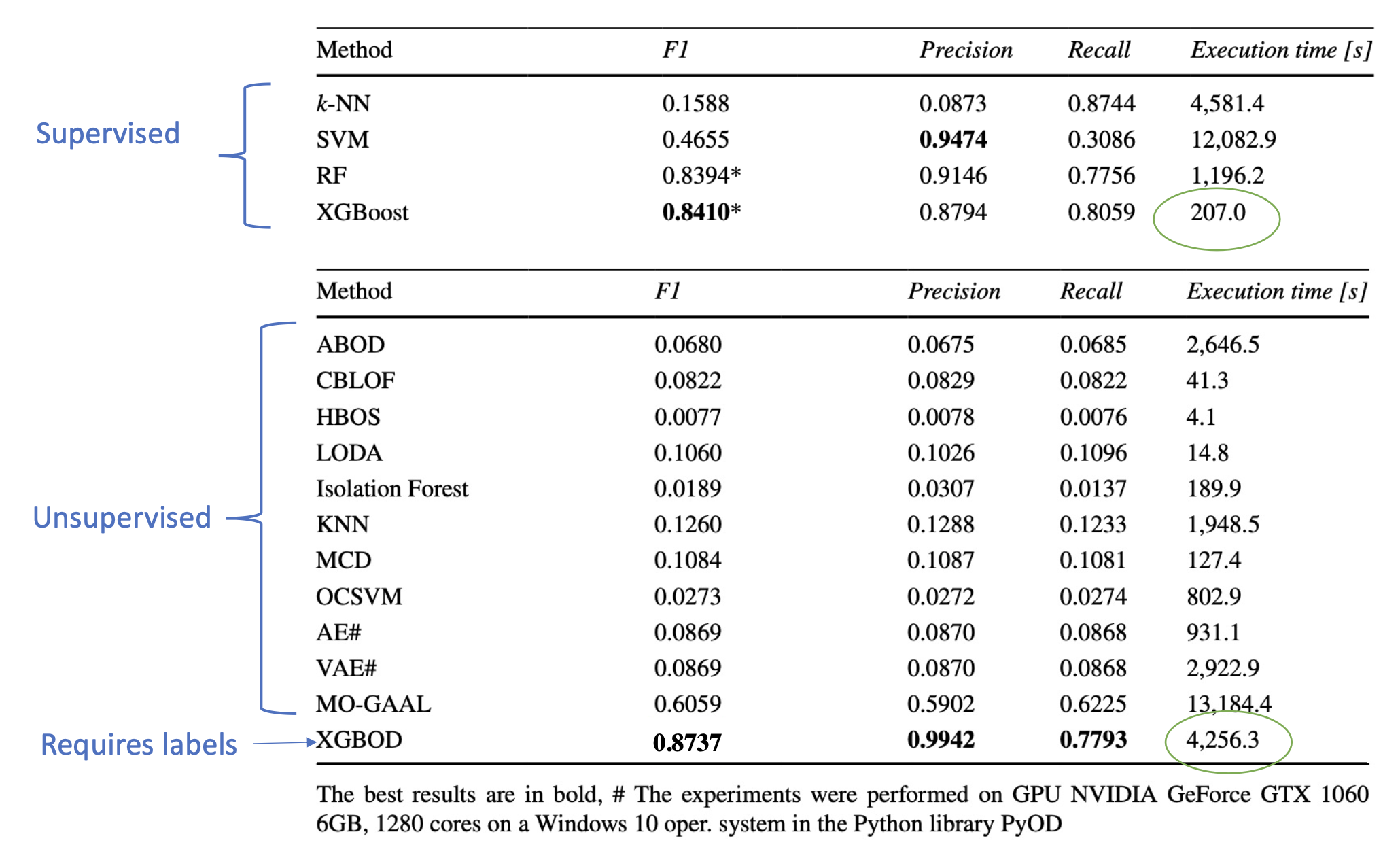}
\end{table}
  
\begin{table}[]
\caption{Comparison of tree boosting systems considering several characteristics. Taken from \protect\cite{chen2016xgboost}.} \label{f:s1}
\includegraphics[width=1\textwidth]{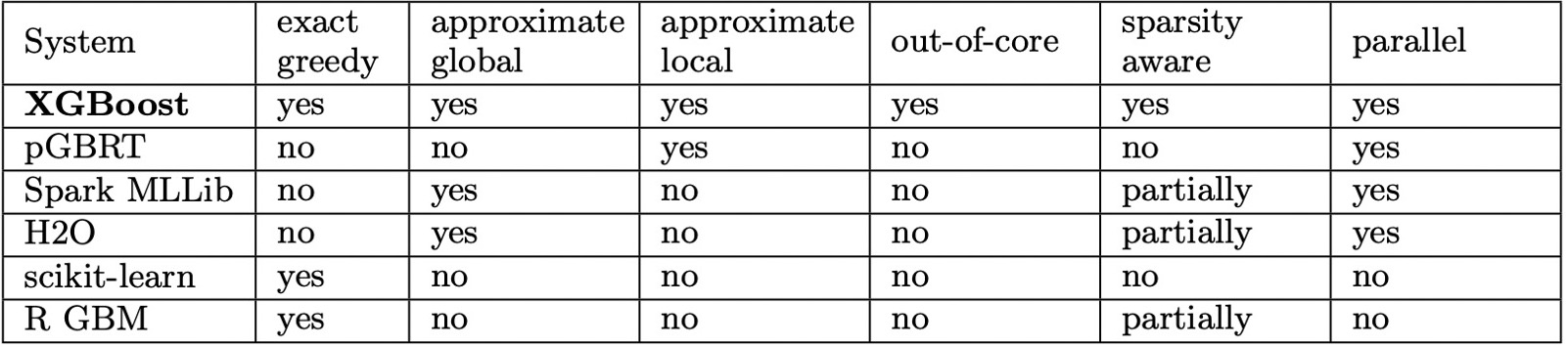}
\end{table}

\section{Experiments} \label{s:experiments}

Two experiments are presented in this section for private data (non-synthetic) containing more than 150 features.

\subsection{Experiment 1}
The first experiment aims at studying XGBoost performance in relation to dataset size and class distribution. In addition, it aims at understanding the impact of parameter fine-tuning with random search, given that random search is more efficient than grid search \cite{bergstra2012random}.
 
\subsubsection{Datasets.}
\begin{figure}[]
\caption{Initial datasets created for Experiment 1.} \label{f:e1}
\includegraphics[width=1\textwidth]{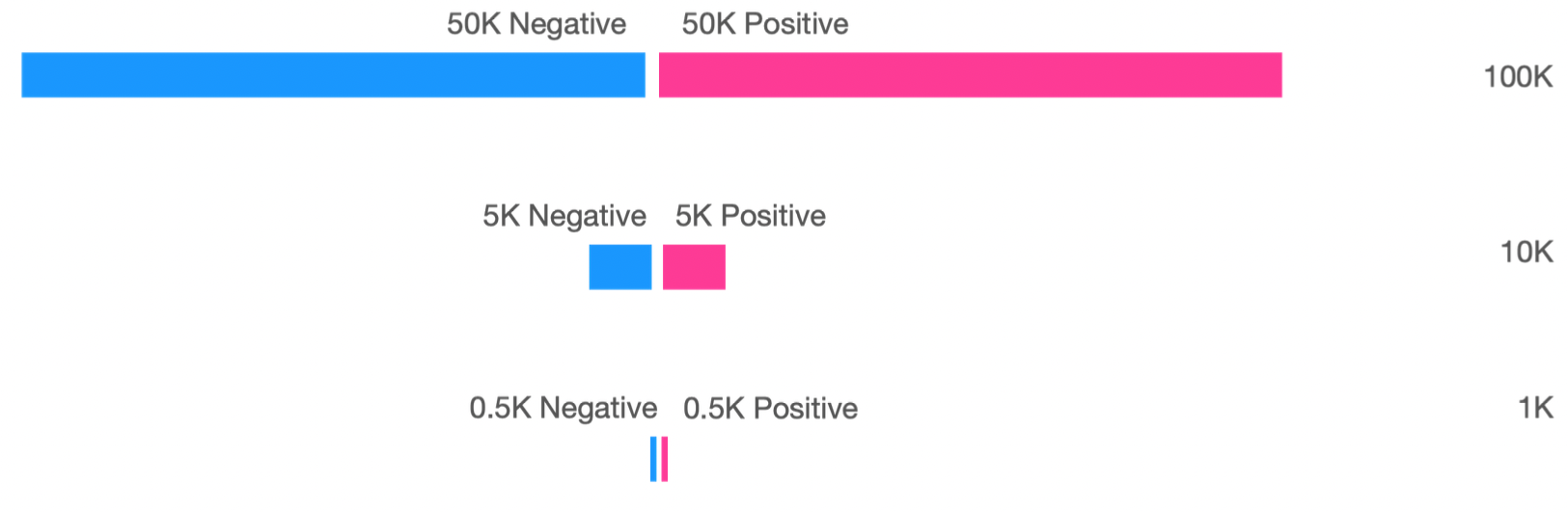}
\end{figure}

\begin{figure}[]
\caption{Datasets created from the partition shown in Figure \ref{f:e1}, for 100K, 10K and 1K  samples. } \label{f:e1b}
\includegraphics[width=1\textwidth]{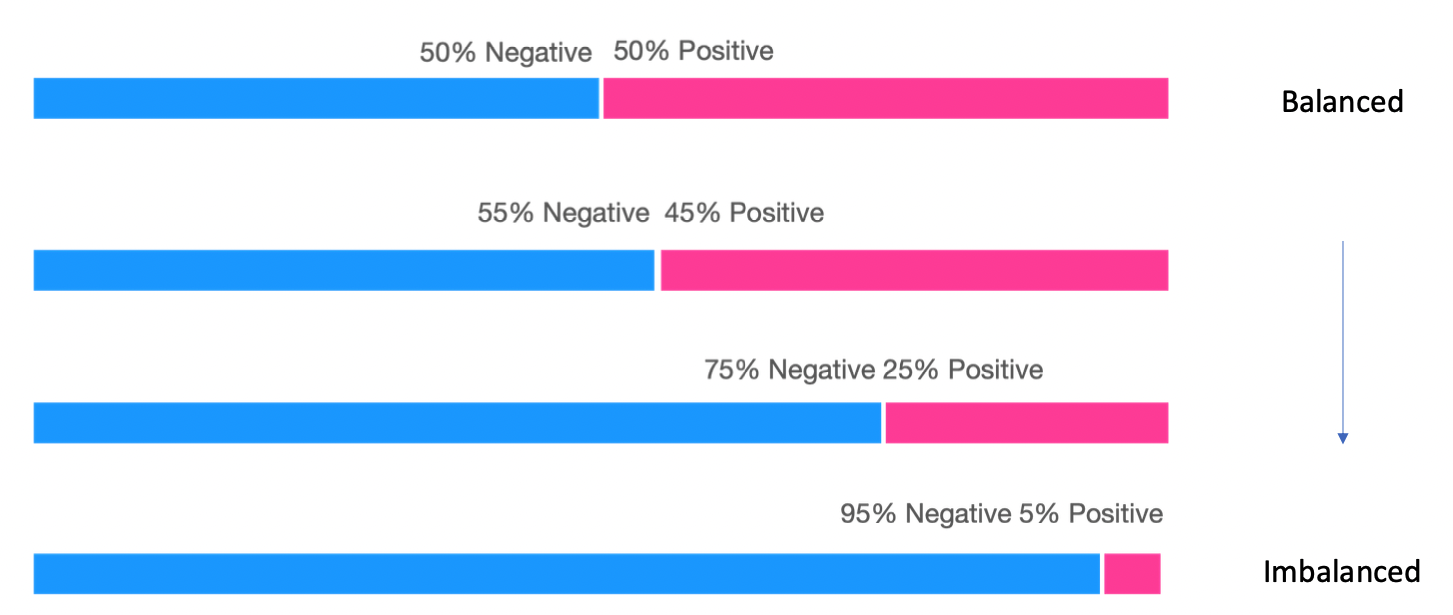}
\end{figure}

Three datasets were created with 100, 10 and 1 thousand (K) samples, respectively, with an equal number of negative and positive samples in each dataset, as illustrated in Figure \ref{f:e1}. Then each subset was sampled such that it had four distributions from balanced to highly imbalanced (N\%-P\%), such that the positive class was under-sampled from: 50\%-50\%, 55\%-45\%, 75\%-25\%, to 95\%-5\%. See Figure \ref{f:e1b}.

\subsubsection{Pipeline}
The datasets were partitioned using 80-20\% for training-testing. Data was prepared as follows:
\begin{itemize}
\item Numerical data was scaled between 0 and 1. 
\item Categorical data was encoded with an ordinal encoder, such that values unseen during training received a reserved value.
\end{itemize}

A Vanilla XGboost, with most default parameters \cite{xgboost_doc}, was tested with the following setup:

\begin{itemize}
\item Binary logistic objective function,
\item Handling of missing values,
\item Random state equal to 100 for reproducibility,
\item Evaluation metric: AUC-PR, 
\item Tree method: GPU histogram
\item Maximum depth of a tree equal to 6,
\item Learning rate equal to 0.3,
\item Subsample ratio of training instances before growing trees equal to 1,
\item Subsampling of columns by tree equal to 1, and 
\item Number of trees equal to 100.
\end{itemize}

In addition, Random Search (RS)-Tuned XGBoost models were obtained in cross-validation over the following space:
 
\begin{itemize}
\item Maximum depth of a tree in values equal to 3, 6, 12, and 20,
\item Learning rate in values equal to 0.02, 0.1, and 0.2,
\item Subsample ratio of training instances before growing trees equal to 0.4, 0.8, and 1,
\item Subsampling of columns by a tree in values equal to 0.4, 0.6, and 1, and 
\item Number of trees equal to 100, 1000, and 5000.
\end{itemize}

This set of parameters were tested over random search, given that a wining Kaggle entry for fraud detection use them \cite{xgboost_tuning}.

\subsection{Results of Experiment 1} \label{s:results}
Table \ref{t:r1} presents the results obtained by the Vanilla XGboost configuration. As expected, XGBoost improves as the dataset increases in size. Focusing on $F_1$ score, we observe that for the balanced distribution and up to 75\%-25\%, there is a larger improvement from 1K to 10K, and a smaller increase from 10K to 100K. In contrast, for the most imbalanced case of 95\%-5\%, the most considerable improvement occurs between 10K and 100K samples. 

\begin{figure}[]
\begin{center}
\caption{F1 scores obtained by Vanilla \mbox{XGBoost} and RS-Tunned XGBoost from Tables \ref{t:r1} and \ref{t:r2}. } \label{f:compa1}
\includegraphics[width=1\textwidth]{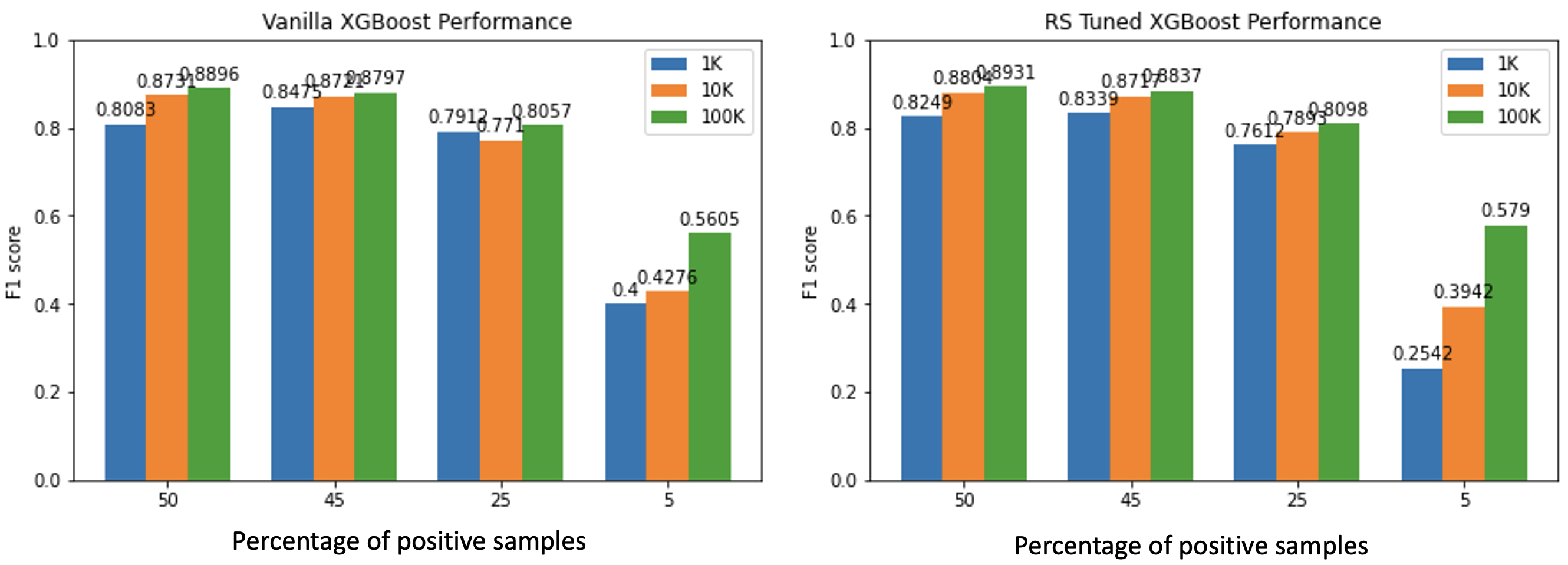}
\end{center}
\end{figure}

Table \ref{t:r2} presents the results obtained by RS-Tuned XGBoost, and Table \ref{t:r3} shows a comparison of $F_1$ scores taken from Table \ref{t:r1} and Table \ref{t:r2}, that is a comparison between Vanilla XGBoost and RS-Tuned XGBoost. Random Search fine-tuning did not present improvement for 1K and 10K samples. Moreover, random search fine-tuning deteriorated the performance for the highly imbalanced scenario of 95\%-5\%. In contrast, the benefit of random search fine-tuning can be observed for the larger dataset of 100K samples. These results indicate that XGBoost default parameters are the recommended choice for small and medium datasets in comparison to the selected parameters tested for RS-Tuned XGBoost. For larger datasets, random search fine-tuning improves detection performance at the cost of being computational expensive. A Summary of the results can be seen in Figure \ref{f:compa1}.

\begin{table}[]
\caption{Performance for Vanilla XGBoost.} \label{t:r1}
\includegraphics[width=1\textwidth]{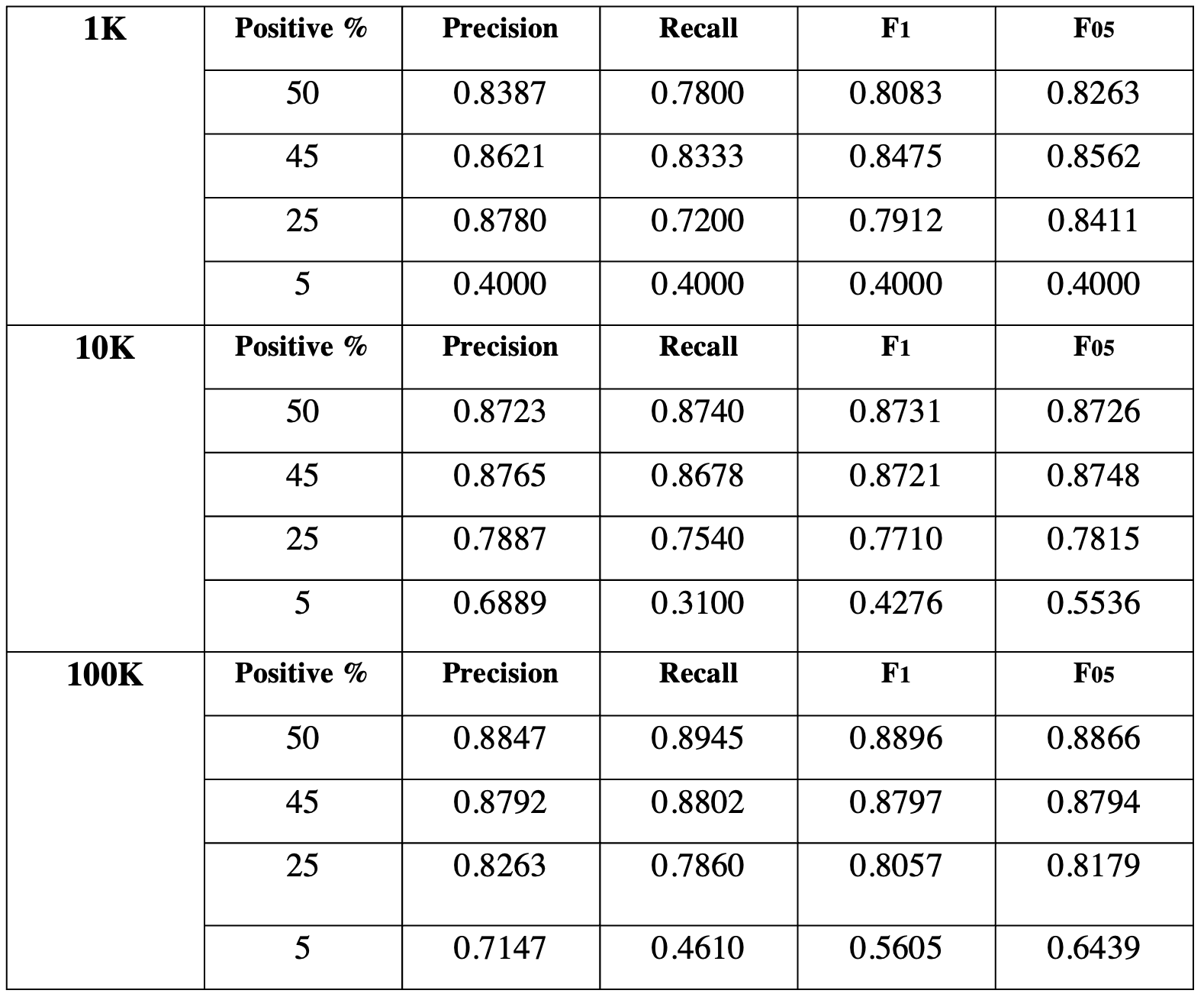}
\end{table}

\begin{table}[]
\caption{Performance for RS-Tuned XGBoost. For 1K, random search run three times. Reported mean values.} \label{t:r2}
\includegraphics[width=1\textwidth]{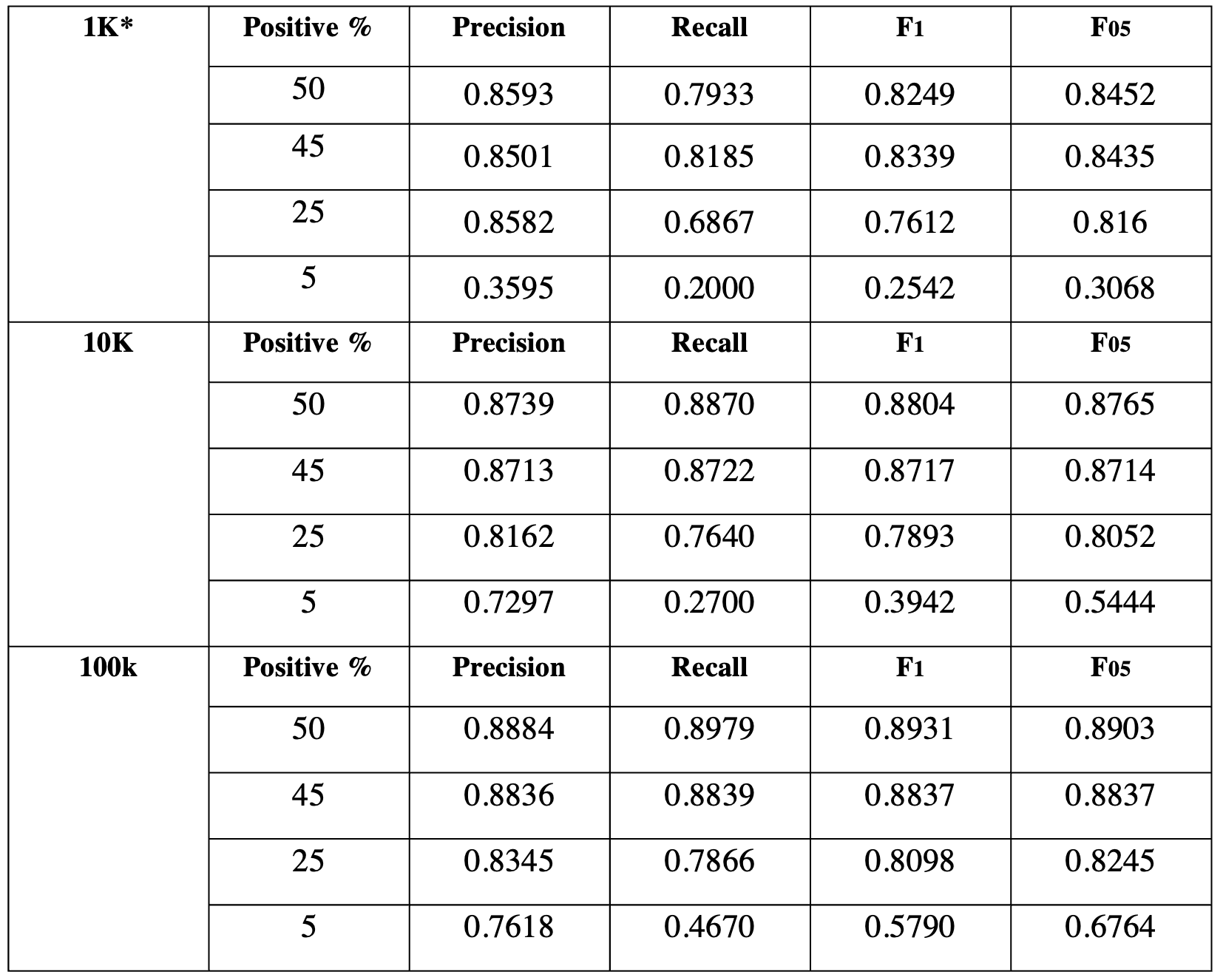}
\end{table}

\begin{table}[]
\begin{center}
\caption{Comparison between Vanilla XGBoost and RS-Tuned XGboost. Best $F_1$ scores in bold. For 1K, random sampling run three times. Fine tuning did not provide consistent improvement for small and medium datasets. In addition, the standard deviation for 95\%-5\% is high. In contrast, fine tuning with random search consistently improve performance for the dataset of 100K samples. } \label{t:r3}
\includegraphics[width=0.7\textwidth]{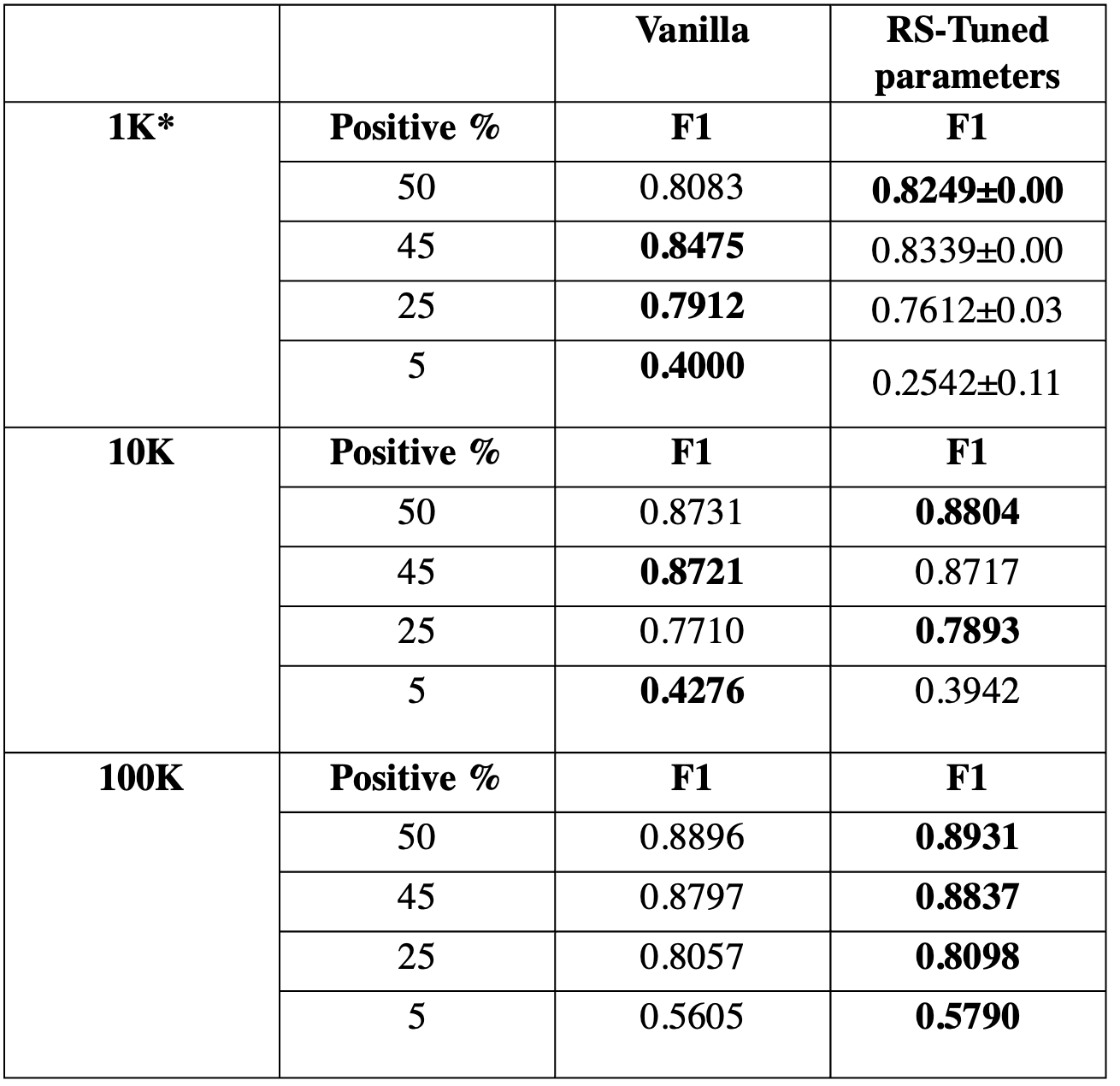}
\end{center}
\end{table}

\begin{figure}[]
\begin{center}
\caption{F1 scores obtained by Vanilla \mbox{XGBoost} and RS-Tunned XGBoost for the dataset of 100K samples, results from Table \ref{t:r3}. } \label{f:compa2}
\includegraphics[width=1\textwidth]{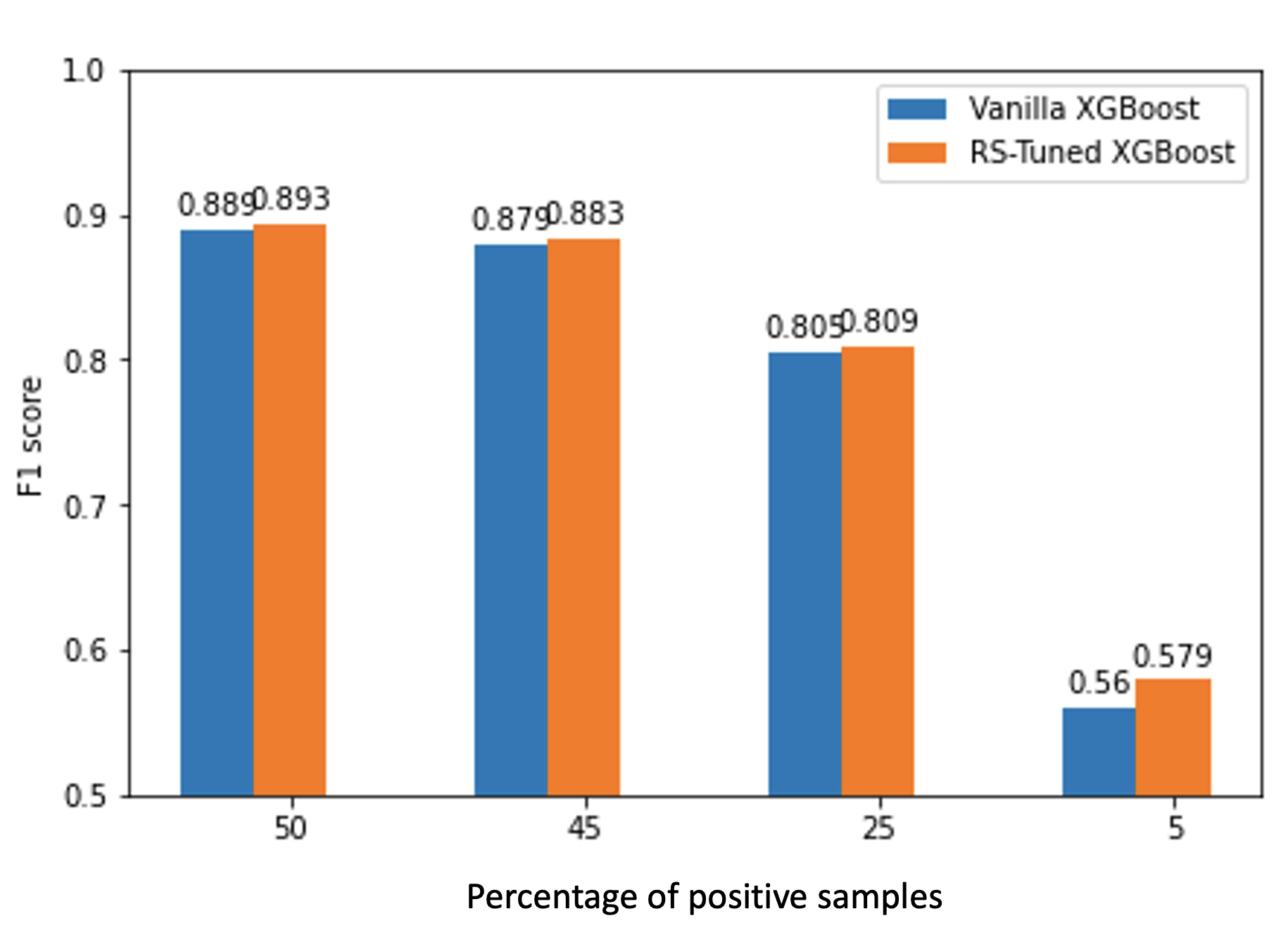}
\end{center}
\end{figure}

\subsection{Experiment 2}
Since XGBoost performs best when dealing with balanced datasets and is strongly affected for the highly imbalanced dataset containing only 5\% positive samples, the second experiment was designed to study the effect of sampling as a technique to improve detection for imbalanced distributions.

\subsubsection{Datasets.}
\begin{figure}[]
\begin{center}
\caption{Data preparation for Experiment 2} \label{f:E2}
\includegraphics[width=1\textwidth]{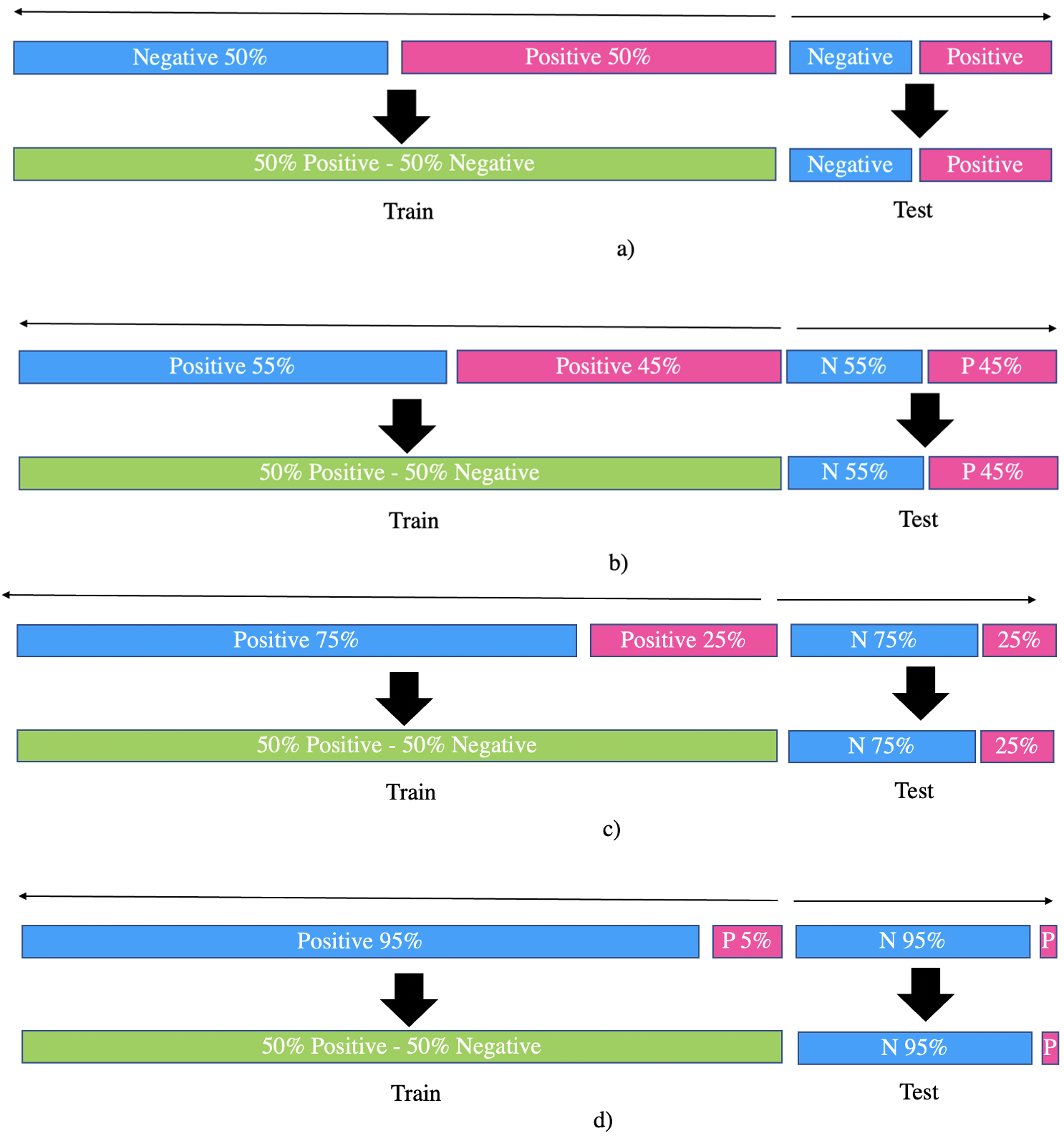}
\end{center}
\end{figure}

Figure \ref{f:E2} illustrates how data was sampled, such that each training set had equal number of positive and negative samples, while the test set reflected four distributions: 50\%-50\%, 55\%-45\%, 75\%-25\%, and 95\%-5\%. A time point was selected to split the data on train and test sets. Then, sampling followed to obtain the desired distribution for each set and partition. This procedure was repeated for 1K and 10K samples, with 80\%-20\%, train-test partition.

\subsection{Results Experiment 2}
Table \ref{t:t4} shows the results when training set is sampled to have equal number of positive and negative samples and XGBoost is tuned using random search. In this case, the estimated $F_1$ becomes unrealistic for distributions of 75\%-25\% and 95\%-5\%. Besides, recall improves but precision worsens. Since we observed that randoms search fine tuning did not provide consistent improvement for datasets of 1K and 10K, a vanilla XGBoost was also tested. The comparison between Vanilla and RS-Tuned XGBoost can be seen in Table \ref{t:t5}. Sampling on training set did not produce consistent improvement on Vanilla or RS-Tuned XGBoost.

\begin{table}[]
\begin{center}
\caption{Effect of sampling to 50\%-50\% (Negative - Positive) on training set for datasets of size 1K and 10K. For RS-Tuned XGBoost, recall improves but precision worsens. The estimated $F1$ score becomes unrealistic for 75\%-25\% and 95\%-5\% distributions. } \label{t:t4}
\includegraphics[width=0.9\textwidth]{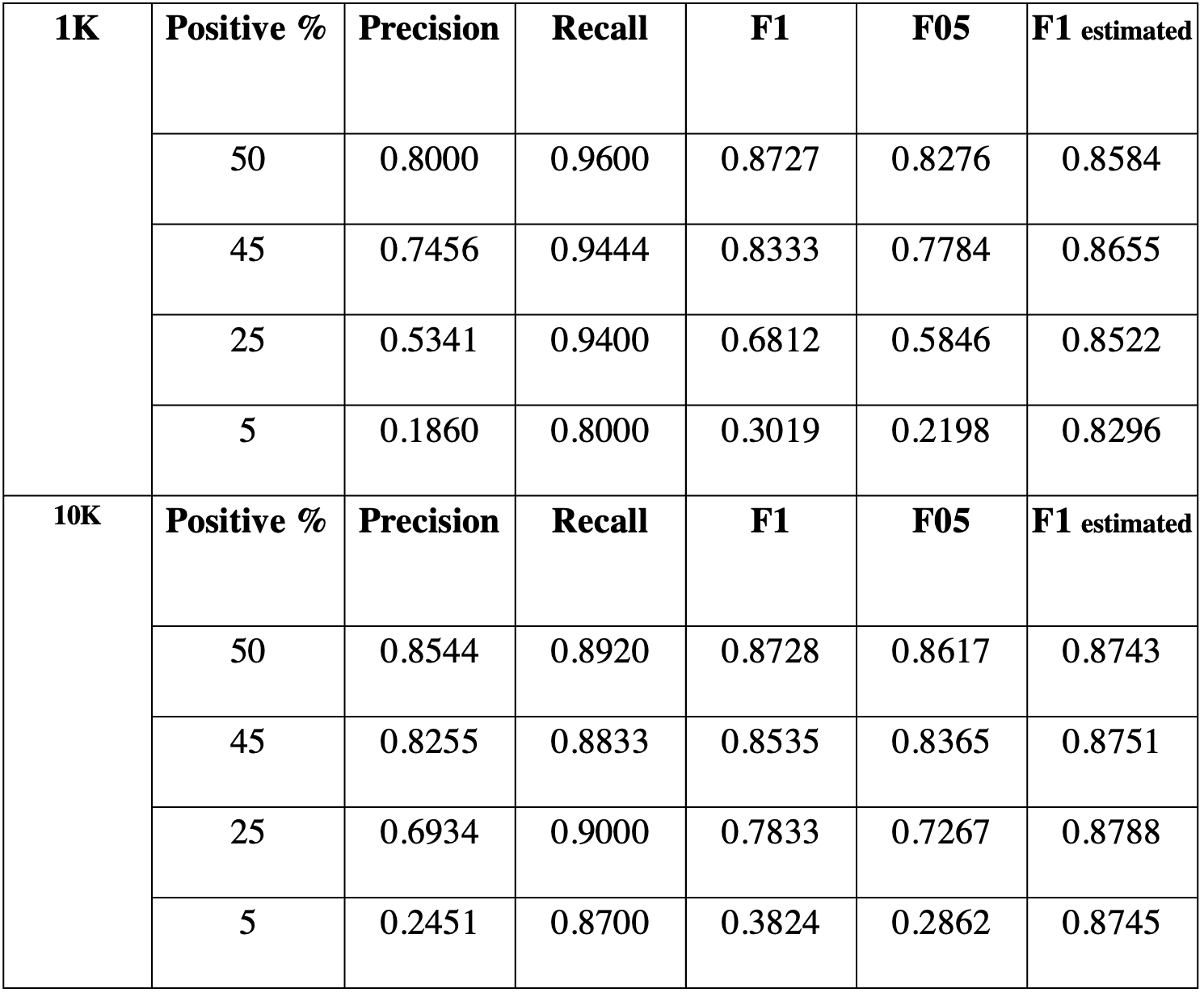}
\end{center}
\end{table}

\begin{table}[]
\begin{center}
\caption{RS-Tuned XGboost and Vanilla XGboost under the effect of sampling to 50\%-50\% (Negative - Positive) on training set for datasets of size 1K and 10K. There is no consistent improvement from using any of the configurations.} \label{t:t5}
\includegraphics[width=0.8\textwidth]{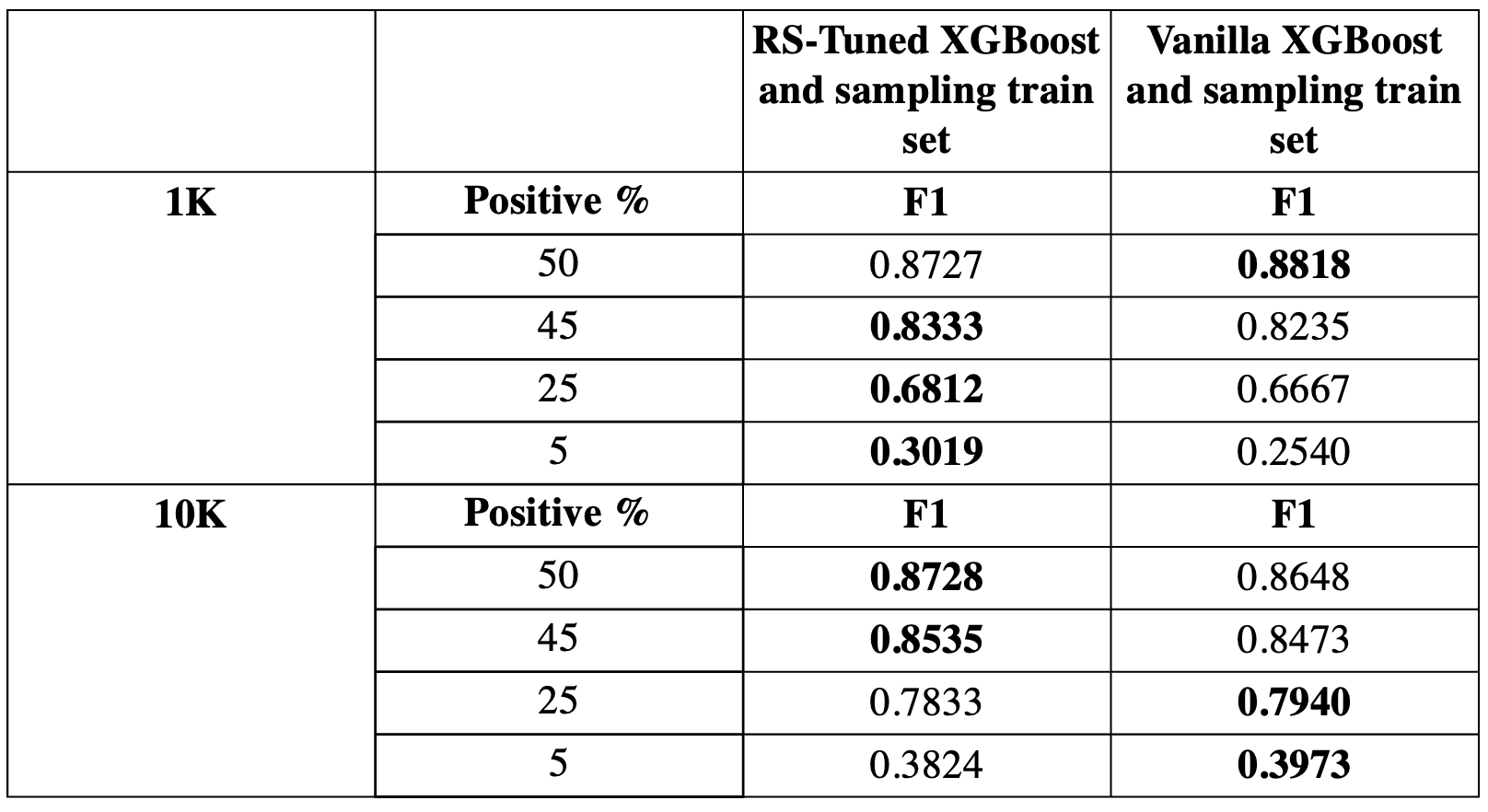}
\end{center}
\end{table}


\section{Conclusions} \label{s:conclusions}
This work focused on evaluating XGBoost on balanced and imbalanced datasets. It explained how evaluation measures are to be interpreted in general for detection systems or binary classifiers. The provided examples showed that accuracy is deceiving when datasets become imbalanced. Besides, it showed that while it is essential to observe precision and recall values, $F$ scores summarise performance. Depending on the application, $F_1$ should be considered if precision and recall are equally weighted, $F_{0.5}$ if precision is preferred over recall, and $F_2$ if recall is more important than precision. 

In addition, this work reviewed XGBoost, which stands out in various benchmarks as a recommended boosting system \cite{chen2016xgboost},\cite{hajek2022fraud}. The proposed pipeline scales numerical values between 0 and 1, and encodes categorical data, giving a reserved value when unseen categories appear in test. Preliminary experiments showed that scaling numerical values does not have an impact for small and medium dataset, but improves performance when a dataset reaches 100K samples. Therefore, scaling numerical values has been used in the pipeline thought out the experiments. Besides, as expected, this report empirically demonstrated that XGboost increases its detection performance as the dataset size increases. Moreover, the experiments showed that XGBoost's performance decreases as the data becomes more imbalanced.

Sampling was tested as a solution to overcome the problem of decreased performance when imbalance increases. Sampling to balance the training set did not provide consistent improvement. Similarly, related work found that random under-sampling deteriorates XGBoost's performance \cite{hajek2022fraud}. There are various methods implemented to deal with the problem of decreased performance for imbalanced datasets \cite{lemaitre2017imbalanced}. These methods will be tested in future work. In addition, forthcoming tests will involve fine-tuning XGBoost weights that observe the ratio between positive and negative samples. 

Finally, future directions include comparing XGBoost against Graphs, autoencoders, and generative adversarial approaches such as Multi-Objective Generative Adversarial Active Learning (MO-GAAL) to deal with the lack of labels.

\section*{Author contributions}
G.V. wrote the paper, created the dataset partitions, developed the pipeline, train and tested XGBoost, and evaluated the results. A.S., S.D. and A.D. provided insights on previous implementations. A.S., K.S., and V.J. collected the dataset for the experiments. K.S. started initial experiments with graphs. 
 
 \section*{Acknowledgment}

We would like thank Michael Weichert for his feedback on drafts of this report, and Rafael Niegoth for providing business knowledge and assigning the task of developing and evaluating a detection system. We would like thank Praveen Maurya and Steffen Wenzel for their support with the servers. In addition, we would like to thank the NVIDIA reviewers and organisers Lilac Ilan, Krystian Garbaciak, Melanie Mangum, and Bridget Johnson for their feedback and support. 

 \bibliographystyle{splncs04}
 \bibliography{References.bib}
\end{document}